\documentclass[11pt]{article}

\usepackage[final]{acl}

\usepackage{times,latexsym}
\IfFileExists{pcrr8t.tfm}{}{}
\IfFileExists{phvr8t.tfm}{}{}
\usepackage[T1]{fontenc}\usepackage[utf8]{inputenc}
\usepackage{microtype,upquote}    % straight backticks/quotes in \texttt and \verb
\IfFileExists{inconsolata.sty}{\usepackage{inconsolata}}{}
\usepackage{graphicx,xcolor,booktabs,amsmath,amssymb,algorithm,algpseudocode,tikz}
\usetikzlibrary{positioning, arrows.meta, calc, fit, backgrounds, shapes.geometric}

\title{Intervention-Certified Amplification:\\ A Validity Gate for $\sigma$-Weighted Post-Training}
\author{
Leiyue Chen$^{*}$,
Meng Wei$^{*}$,
Xuefeng Hu,
Wei Liu$^{\dagger}$
\\
ByteDance
\\
\texttt{\{leiyue.ml, xmechira, liujikun63\}@gmail.com,\{xuefeng.hu\}@bytedance.com}
}
\begin{document}
\maketitle
\begingroup
\renewcommand{\thefootnote}{} % 关掉自动数字编号
\footnotetext{\textsuperscript{*} Equal contribution.}
\footnotetext{\textsuperscript{$\dagger$} Corresponding author.}
\endgroup

\begin{abstract}
Hybrid post-training often mixes supervised fine-tuning (SFT) and reinforcement learning (RL), using training-side signals to set each sample's balance. Such signals characterize the anchor data and policy, but not whether the emphasized behavior helps in the target task and evaluation setting. This mismatch can reactivate and amplify pretrained behavior even after its literal form is removed from supervision. We demonstrate this with \textbf{DIAA} (Dual-Implicit Adversarial Amplification), a per-sample $\sigma$-weighted SFT/RL hybrid. With all literal Python fences removed from its anchor, DIAA emits unexecuted Tool-Integrated Reasoning (TIR) on $29\%$ of MATH-500 prompts, three times vanilla GRPO. Paired interventions show $10.6$ percentage points of accuracy recovery when TIR is suppressed and comparable contributions from SFT imitation and $\sigma$-amplification; a student trained on a $0\%$-TIR teacher anchor emits it at $45.8\%$. \textbf{Intervention-Certified Amplification (ICA)} gates $\sigma$ with paired-rollout target evidence. Frozen and online ICA reduce TIR to $13.8\%$ and $11.2\%$, halve residual drag, improve accuracy from $47.5\%$ to $48.7\%$ and $49.3\%$, and transfer to held-out GSM8K without retuning.
\end{abstract}

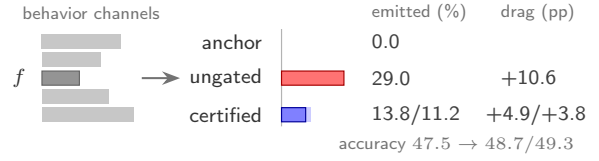
\begin{figure}[t]\centering
\resizebox{\columnwidth}{!}{%
\begin{tikzpicture}[
  font=\sffamily, >=Stealth,
  hd/.style={font=\sffamily\tiny, text=black!55},
  rl/.style={font=\sffamily\scriptsize, text=black!80, anchor=east},
  vl/.style={font=\sffamily\scriptsize, text=black!75, anchor=west},
]
\useasboundingbox (0.02,-0.50) rectangle (7.02,1.92);

\node[font=\sffamily\small\bfseries, text=black!85] at (3.52,1.72)
  {Certify a behavior before amplifying it};

% ── left: conceptual repertoire, deliberately unquantified ──────────────
\node[hd, anchor=west] at (0.05,1.25) {behavior channels};
\foreach \y/\len in {0.90/0.95, 0.68/0.70, 0.24/0.80, 0.02/1.10}
  \fill[black!22] (0.42,\y-0.085) rectangle (0.42+\len,\y+0.085);
\fill[black!42, draw=black!60, line width=0.4pt] (0.42,0.375) rectangle (0.87,0.545);
\node[font=\sffamily\scriptsize, anchor=east, text=black!80] at (0.36,0.46) {$f$};

\draw[->, line width=0.7pt, draw=black!55] (1.62,0.46) -- (2.08,0.46);

% ── right: the one instantiated channel, all values measured ────────────
\node[hd, anchor=west] at (4.25,1.25) {emitted (\%)};
\node[hd, anchor=west] at (5.80,1.25) {drag (pp)};
\draw[black!30, line width=0.4pt] (3.30,-0.12) -- (3.30,1.05);

\node[rl] at (3.20,0.90) {anchor};    \node[vl] at (4.25,0.90) {0.0};
\node[rl] at (3.20,0.46) {ungated};   \node[vl] at (4.25,0.46) {29.0};
\node[rl] at (3.20,0.02) {certified}; \node[vl] at (4.25,0.02) {13.8/11.2};
\node[vl] at (5.80,0.46) {+10.6};
\node[vl] at (5.62,0.02) {+4.9/+3.8};

\fill[red!55, draw=red!65!black, line width=0.4pt] (3.30,0.375) rectangle (4.05,0.545);
\fill[blue!20] (3.30,-0.065) rectangle (3.657,0.105);
\fill[blue!50, draw=blue!60!black, line width=0.4pt] (3.30,-0.065) rectangle (3.590,0.105);

\node[hd, anchor=east] at (6.95,-0.34) {accuracy $47.5 \rightarrow 48.7/49.3$};
\end{tikzpicture}}
\caption{\textbf{Certify a behavior before amplifying it.} $\sigma$-weighted post-training selects one channel out of the pretrained behavior repertoire and amplifies it; an intervention-derived certificate gates that channel. The channel column on the left is a conceptual depiction and carries no measurement; only the highlighted channel $f$ is instantiated experimentally, as unexecuted Python-fence form (TIR). \emph{Paired drag} is the accuracy recovered on matched prompts when an intervention removes $f$; after gating it is residual, so lower is better. Certified values are the frozen and online instantiations of the same gate rather than a ranking; Tab.~\ref{tab:experiments} gives the full comparison, including the ungated baseline and non-certificate controls.}\label{fig:teaser}
\end{figure}

\section{Introduction}\label{sec:intro}

When should a model imitate a demonstration, and when should it explore on its own? Recent post-training methods answer it with training-side signals. OPD \citep{opd2024,rethinkopd2026} densifies teacher supervision, \citet{rlsd2026} densify environment feedback through self-distillation, BREAD \citep{bread} splices expert-anchored prefixes, AESL \citep{aesl} tunes a lightweight SFT cold start, and RAFT \citep{raft} and STaR \citep{star} keep reward-filtered or self-generated rationales. However the signal is formed, it is computed from training-side evidence; whether the behavior it emphasizes helps in the target task and evaluation setting is not itself tested.

\paragraph{A concrete failure mode.}
This gap is testable, and data cleaning does not close it. After deleting every literal Python code fence from the anchor used to post-train Qwen2.5-1.5B, the model still emits unexecuted Tool-Integrated Reasoning (TIR) on $29\%$ of MATH-500 prompts, three times vanilla GRPO. Where suppressing the form succeeds, it recovers $+10.6$\,pp of accuracy. Because the behavior is absent from the anchor yet reappears, post-training reactivates a pretrained form rather than literally imitating the RFT. Training-side absence therefore does not certify target-side usefulness.

\paragraph{Intervention-based diagnosis.}
Our probe is \textbf{DIAA} (Sec.~\ref{sec:diaa}), a per-sample $\sigma$-weighted SFT/RL hybrid derived from the Bayes-optimal anchor--policy discriminator and used to isolate amplification rather than to claim accuracy. Each checkpoint is evaluated under matched baseline and positive-framing prompts. On the subset where the intervention removes TIR, accuracy recovery rises along GRPO~$\to$~SFT~$\to$~DIAA as $+2.2 \to +8.5 \to +10.6$\,pp. We call this recovered gap \emph{paired accuracy drag}: it is the accuracy regained when the behavior disappears on the same prompt, while the never-TIR cell estimates the intervention's generic effect. After that control, the pathway splits into comparable SFT-imitation ($+4.7$\,pp) and $\sigma$-amplification ($+5.3$\,pp) components. The label is operational, not a claim that TIR is intrinsically harmful.

\paragraph{Certify before amplifying.}
ICA gates the original weight with a certify token, $\sigma_{\text{eff}}(x)=\sigma(x)c(x;\theta_c)$. The frozen instantiation holds $c$ from a prior paired probe fixed; the online instantiation updates $\theta_c$ from paired-rollout deltas. ICA acts only at the per-sample weight, preserving the loss family and anchor source, and reduces to DIAA when $c\equiv1$.

\paragraph{Roadmap.} Sec.~\ref{sec:background}--\ref{sec:setup} present DIAA and the observation; Sec.~\ref{sec:paired} diagnoses drag and its granularity signature; Sec.~\ref{sec:method} defines ICA; Sec.~\ref{sec:experiments} evaluates frozen and online gates against non-certificate controls.

\section{Background and DIAA}\label{sec:background}

\paragraph{Terminology.} \textbf{Rejection-sampling fine-tuning (RFT)} in the RAFT \citep{raft} / STaR \citep{star} lineage is a reward-filtered dataset; here reward-$1.0$ samples form the anchor. A \textbf{Python fence} is the Markdown delimiter \verb|```python| that opens a code block. \textbf{Tool-Integrated Reasoning (TIR) form} denotes reasoning presented inside such a fence together with a claimed \texttt{output} block, with no actual execution. \textbf{Cargo-cult drag} is the accuracy recovered on the rescued paired cell when the intervention removes TIR form; the certificate additionally subtracts a never-TIR control to absorb generic suffix effects. \textbf{$\sigma$-weighted SFT/RL} denotes $\sigma(x)\mathcal{L}_{\text{SFT}} + (1-\sigma(x))\mathcal{L}_{\text{RL}}$ with $\sigma(x)\in[0,1]$: no sample is routed exclusively to one branch, and high $\sigma$ favors imitation. A \textbf{paired-intervention probe} evaluates the same checkpoint under suffixes $\tau_{\text{base}},\tau_{\text{int}}$; within-prompt differences isolate the suffix's effect on a candidate trace feature. \textbf{ICA} is the intervention-derived gate $c$ that modifies the per-sample weight as $\sigma_{\text{eff}}=\sigma c$, with a frozen and an online implementation.

\paragraph{Filter-by-reward self-RFT.} We sample $K$ rollouts per prompt, score them with a verifier, and keep top-reward traces as anchor RFT. This is sample-efficient but inherits verifier failures: a trace can score reward $1.0$ by chance correlation between an unverified intermediate step and the final answer. With Qwen2.5-1.5B base + math RFT, $26.5$\% of reward-$1.0$ traces contain \verb|```python ... ```output| blocks without actual execution.

\paragraph{DIAA.}\label{sec:diaa}
DIAA's per-sample weight is the inverted Bayes-optimal discriminator between the RFT anchor $\pi_{\text{anchor}}$ and the policy $\pi_\theta$, giving the loss
\begin{equation}\label{eq:diaa-main}
\begin{aligned}
\mathcal{L}
&= \sigma(x)\,\mathcal{L}_{\text{SFT}}
 + (1-\sigma(x))\,\mathcal{L}_{\text{RL}}, \\
\sigma(x)
&= \mathrm{sigmoid}\!\left(-\beta\,\Delta(x)\right), \\
\Delta(x)
&= \log p_\theta(\mathrm{rft})-\log p_\theta(\mathrm{roll}).
\end{aligned}
\end{equation}
The full derivation is in App.~\ref{app:diaa-deriv}. Because $\Delta$ is computed from the current policy's log-probabilities of one observed anchor trace and one on-policy rollout, DIAA trains and queries no separate discriminator network. DIAA admits a per-prompt variant ($\sigma$ shared across rollouts) and a truly-per-sample variant ($\sigma$ per rollout); the choice has measurable consequences (Sec.~\ref{sec:paired}). Two-time-scale convergence is established, conditional on its stated assumptions, in App.~\ref{app:diaa-ttsa}.

\paragraph{TIR and tool grounding.} \citet{linxu2025tir} formalize TIR's strict-expansion property under a deterministic external oracle. Without such an oracle, TIR form does not necessarily indicate tool capability: a $\sigma$-weighted procedure can amplify the surface form independent of whether the model can execute the code. DIAA is therefore an instrument for measuring amplification, not a claim about tool use.

\section{Setup and Cross-Sectional Observation}\label{sec:setup}

\paragraph{Models and data.} Qwen2.5-1.5B base \citep{qwen25} with LoRA following TINA's \citep{tina} LoRA recipe; RFT is $94{,}865$ self-generated samples on GSM8K+MATH train, filtered by boxed-answer match. Checkpoints (Tab.~\ref{tab:chkpt}): vanilla GRPO; pure SFT on the same RFT; DIAA truly-per-sample on raw RFT ($26.5$\% TIR contamination); DIAA truly-per-sample on RFT with literal Python code fences removed; pure SFT on a teacher-generated cross-model anchor (DeepSeek-Math-7B-Instruct \citep{deepseekmath} teacher rollout, $0$\% TIR in the RFT).

\begin{table}[t]\centering\small
\setlength{\tabcolsep}{4pt}
\begin{tabular}{@{}lrrr@{}}\toprule
Checkpoint        & Acc  & TIR  & RFT-TIR \\
\midrule
vanilla GRPO      & 44.4 & 9.2  & n/a  \\ SFT (self)        & 40.6 & 17.2 & 26.5 \\
DIAA (raw)        & 41.2 & 44.2 & 26.5 \\ DIAA (clean)      & 47.5 & 29.0 & 0    \\
SFT (teacher)     & 37.6 & 45.8 & 0    \\
\bottomrule\end{tabular}
\caption{MATH-500 triangulation (all columns in \%). DIAA amplifies TIR even after literal Python fences are removed; a clean teacher RFT ($0\%$ TIR) still yields $45.8\%$ student TIR, indicating pretrained-channel reactivation rather than literal RFT imitation (App.~\ref{app:ceiling-limitations}, C1).}
\label{tab:chkpt}\end{table}

DIAA on clean RFT outperforms vanilla GRPO by $+3.1$\,pp, yet TIR persists across SFT-touched checkpoints. Because cross-sectional $\alpha_{\text{text}}-\alpha_{\text{TIR}}$ conflates selection with causality, a lower TIR cross-section would not by itself certify a better method and a higher one would not prove harm. The paired cells below are the unit of causal evidence.

\section{Empirical Evidence of Causal Drag}\label{sec:paired}

\paragraph{Design.} For each prompt $q$ and checkpoint, we compare $y_{\text{base}}=M(q,\tau_{\text{base}})$ and $y_{\text{int}}=M(q,\tau_{\text{int}})$, where $\tau_{\text{int}}$ is positive-framing and avoids forbidden-token priming (App.~\ref{app:prompts}). The rescued cell $(T,N)$ gives a controlled within-prompt contrast for suppressing TIR at fixed prompt difficulty; the never-TIR cell $(N,N)$ bounds the generic effect of the suffix itself.

\paragraph{Main paired result.}\label{sec:paired-main}

\begin{table*}[t]\centering\small
\begin{tabular}{lcccccc}\toprule
& \multicolumn{3}{c}{Positive intervention $\tau_+$} & \multicolumn{3}{c}{Generic intervention $\tau_g$} \\
\cmidrule(lr){2-4}\cmidrule(lr){5-7}
Checkpoint & $(T,N)\ n$ & $\Delta_{TN}$ & $\Delta_{NN}$ ctrl & $(T,N)\ n$ & $\Delta_{TN}$ & $\Delta_{NN}$ ctrl \\
\midrule
$R_{03}$  vanilla GRPO            & 45  & $+2.2$\,pp  & $+0.4$\,pp  & 44 & $+2.3$\,pp & $+0.2$\,pp \\ $R_{15}$  pure SFT (self)         & 71  & $+8.5$\,pp  & $+2.0$\,pp  & 59  & $+8.5$\,pp & $+3.9$\,pp \\
$R_{14v3}$ DIAA, raw              & 140 & $+10.7$\,pp & $-2.2$\,pp & 100 & $+8.0$\,pp & $+0.0$\,pp \\ $R_{14v4}$ DIAA, clean            & 113 & $+10.6$\,pp & $-1.2$\,pp & 91  & $+13.2$\,pp & $-0.6$\,pp \\
$R_{15b}$ pure SFT (teacher)      & 170 & $+7.6$\,pp  & $-0.4$\,pp & 116 & $+10.3$\,pp & $+1.9$\,pp \\
\bottomrule\end{tabular}
\caption{Paired rescue, read \emph{diagnostically}: $\Delta_{TN}$ is the accuracy recovered on the rescued cell when the intervention removes TIR, and $\Delta_{NN}$ the never-TIR control, so a \emph{larger} $\Delta_{TN}$ means more removable drag is present. $\hat C_\phi=\Delta_{TN}-\Delta_{NN}$ rises from $+1.8$\,pp (GRPO) to $+6.5$\,pp (SFT) to $+11.8$\,pp (DIAA), decomposing drag into comparable SFT-contamination and $\sigma$-amplification contributions. The clean teacher anchor also gives positive/generic certificates ($+8.1/+8.5$\,pp), which reduces concern about a purely suffix-specific artifact. After mitigation the same quantity is read in the opposite direction (Sec.~\ref{sec:experiments}).}
\label{tab:paired-main}\end{table*}

Tab.~\ref{tab:paired-main} shows that SFT contamination is intervention-stable ($R_{15}$ has the same $+8.5$\,pp rescued delta under both interventions), the clean teacher anchor repeats the signal under a generic suffix ($+10.3$\,pp in $TN$ versus $+1.9$\,pp in $NN$), and SFT and $\sigma$ contributions are comparable. Thus interventions targeting only the $\sigma$ branch can recover at most part of the observed cargo-cult drag. The two suffixes agree in direction and magnitude on $R_{14v4}$ as well ($+10.6$ versus $+13.2$\,pp rescued, against $-1.2$ and $-0.6$\,pp controls), which reduces generic prompt confounding but neither eliminates suffix dependence nor shows that TIR is the only affected behavior.

\paragraph{Granularity and subject heterogeneity.} Truly-per-sample $\sigma$ shows $1.5{-}1.7{\times}$ greater intervention persistence under the tested suffixes than per-prompt aggregation. This is consistent with the granularity dial modulating how prompt-sensitive the learned shortcut is. The certificate sign is subject-dependent: on Counting \& Probability, TIR encodes genuine \texttt{math.comb} capability and the paired delta flips sign, motivating subject-conditional fallback in ICA. Detailed cell counts and the per-subject breakdown are in App.~\ref{app:paired-detail}. The drag is therefore real, decomposable, and unevenly distributed across subjects, which is what a gate must respect.

\section{Intervention-Certified Amplification}\label{sec:method}

\paragraph{Composed loss.} Given a feature indicator $\phi : \mathcal{X} \to \{0,1\}$ identifying a candidate causal-drag trace pattern, ICA introduces a \emph{certify token} $\theta_c$ producing a per-sample gate $c(x; \theta_c) \in [0,1]$. The composed loss is
\begin{equation}\label{eq:ica-loss}
\begin{aligned}
\mathcal{L}_{\text{ICA}}
&= \sigma_{\text{eff}}(x)\,\mathcal{L}_{\text{SFT}}
 + (1-\sigma_{\text{eff}}(x))\,\mathcal{L}_{\text{RL}}, \\
\sigma_{\text{eff}}(x)
&= \sigma(x)\cdot c(x;\theta_c).
\end{aligned}
\end{equation}
Eq.~\eqref{eq:ica-loss} reduces to vanilla DIAA when $c \equiv 1$ and shifts converged feature prevalence by $\propto -\mathrm{Cov}_{\pi_{\text{anchor}}}(\sigma c, \phi) / \mathbb{E}[\sigma c]$ (App.~\ref{app:ceiling}, Corollary).

\paragraph{Two instantiations of one gate.} Eq.~\eqref{eq:ica-loss} is the shared principle; it admits two instantiations that differ only in how $c$ is obtained. In the \emph{frozen} instantiation ($R_{20}$), $c$ is computed once before training from a paired-intervention probe on a prior checkpoint and held fixed (App.~\ref{app:prompts}, Alg.~\ref{alg:cert-instance}). In the \emph{online} instantiation ($R_{\text{ICA}}$), a certify token is updated in-loop from paired rollouts (Alg.~\ref{alg:ica}). We treat the online instantiation as our main adaptive implementation and the frozen one as the control that isolates the certificate content from the update rule.

\paragraph{In-loop certify-token update.} At a certificate event, the same minibatch is rolled out under $\tau_{\text{base}}$ and $\tau_{\text{int}}$. The paired delta on $\phi$-activating prompts provides a value-function-implicit signal that updates $\theta_c$, which modulates $c(x;\theta_c)$ for subsequent steps. Certificate adaptation runs on a slow timescale relative to the policy update (App.~\ref{app:diaa-ttsa}); the corresponding fixed point is what motivates using $c$ as an amplification certificate (App.~\ref{app:ceiling-corollary}). Both instantiations therefore share one principle and differ only in when the certificate is read.

\begin{algorithm}[t]\small
\caption{ICA with certify token in DIAA paradigm}\label{alg:ica}\begin{algorithmic}[1]
\Require RFT $\mathcal{D}$, feature $\phi$, suffixes $\tau_{\text{base}}, \tau_{\text{int}}$, certify-token init $\theta_c^{(0)}$, certificate interval $K_c$
\For{step $t = 1, \dots, T$}
  \State Sample minibatch $\mathcal{B} \subset \mathcal{D}$; roll out under $\tau_{\text{base}}$
  \If{$t \bmod K_c = 0$}
    \State Roll out a probe minibatch under $\tau_{\text{int}}$
    \State Compute paired $\Delta$acc on $\phi$-activating prompts $\to$ signal $s_t$
    \State Update $\theta_c \gets \theta_c + \eta_c\,\nabla_{\theta_c} s_t$
  \EndIf
  \State Compute $\sigma_{\text{eff}}(x) = \sigma(x) \cdot c(x; \theta_c)$ for $x \in \mathcal{B}$
  \State Update policy $\theta$ on $\mathcal{L}_{\text{ICA}}$ (Eq.~\eqref{eq:ica-loss})
\EndFor
\end{algorithmic}\end{algorithm}

\section{Experiments}\label{sec:experiments}

\paragraph{Main result.} All runs share the configuration of App.~\ref{app:hparams} and are evaluated on MATH-500 at step $1200$. We accept ICA as a \emph{weak win} when accuracy gains at least $1$\,pp over ungated $R_{14v4}$, the paired $\Delta_{TN}$ at least halves, and Counting \& Probability does not regress by more than $2$\,pp. The subject-conditional fallback of App.~\ref{app:prompts} leaves Counting \& Probability ungated by construction, since its certificate is non-positive there. Both certificate-gated configurations clear the accuracy and residual-drag criteria (Tab.~\ref{tab:experiments}), gaining $+1.2$ and $+1.8$\,pp accuracy over $R_{14v4}$ while more than halving residual drag. The controls separate certificate \emph{content} from mere downweighting. A shuffled certificate and a length heuristic both reduce TIR and residual drag, yet land \emph{below} $R_{14v4}$ in accuracy, and hard-filtering the flagged RFT samples instead of downweighting them is worse still on all three columns. The evidence therefore supports intervention-certified gating as such. The frozen--online gap is small, and we do not treat it as establishing an advantage for online adaptation.

\begin{table}[t]
\centering
\small
\setlength{\tabcolsep}{4pt}
\begin{tabular}{@{}llrrr@{}}
\toprule
Run & Role & Acc & TIR & $\Delta_{TN}$ \\
\midrule
$R_{03}$    & GRPO baseline      & 44.4 & 9.2  & +2.2 \\
$R_{14v4}$  & ungated DIAA       & 47.5 & 29.0 & +10.6 \\
\midrule
$R_{22}$    & hard-filtered RFT  & 44.6 & 33.0 & +9.0 \\
$R_{21}$    & shuffled certificate & 46.1 & 24.5 & +8.1 \\
$R_{24}$    & length heuristic   & 46.4 & 23.0 & +7.9 \\
\midrule
$R_{20}$    & \textbf{ICA}, frozen $c$   & 48.7 & 13.8 & +4.9 \\
$R_{\text{ICA}}$ & \textbf{ICA}, online $c$ & \textbf{49.3} & \textbf{11.2} & \textbf{+3.8} \\
\bottomrule
\end{tabular}
\caption{MATH-500 at step $1200$. For $R_{03}$ and $R_{14v4}$ the $\Delta_{TN}$ column is \emph{diagnostic} (drag present); for the remaining rows it is \emph{residual} drag left after the intervention, where lower is better. $R_{22}$ removes flagged RFT samples outright instead of downweighting them; $R_{21}$ rules out generic regularization; $R_{24}$ rules out length-based confounds.}
\label{tab:experiments}
\end{table}

\paragraph{Consistency across runs.} Repeating the three-arm comparison twice more under the same protocol preserves the ordering of Tab.~\ref{tab:experiments} in every run; App.~\ref{app:experiments} gives the three-run mean $\pm$ SD and the per-run differences. These are matched consistency checks rather than a training-variance estimate, so we draw no significance conclusion from the frozen--online difference.

\paragraph{Held-out transfer.} We evaluated the same three arms on the GSM8K \citep{gsm8k} test split with the MATH-500-derived gates unchanged and no retuning. Both gated arms carry the joint improvement to a set that played no part in constructing the certificate (Tab.~\ref{tab:gsm8k}), trading off slightly against each other.

\begin{table}[t]\centering\small
\setlength{\tabcolsep}{4pt}
\begin{tabular}{@{}lrrr@{}}\toprule
Arm & Acc & TIR & $\Delta_{TN}$ \\
\midrule
ungated DIAA        & 67.9 & 24.7 & +12.4 \\
ICA, frozen $c$     & \textbf{70.7} & \textbf{9.3} & +6.4 \\
ICA, online $c$     & 70.4 & 9.8 & \textbf{+6.1} \\
\bottomrule\end{tabular}
\caption{GSM8K test ($n{=}1319$), certificate frozen from MATH-500 and not retuned. $\Delta_{TN}$ is residual drag, where lower is better.}
\label{tab:gsm8k}\end{table}

\paragraph{Certificate cost.} Gating adds one $\tau_{\text{int}}$ probe minibatch every $K_c$ steps and no additional model; App.~\ref{app:experiments} gives the full accounting. Certified gating thus buys lower emission and lower residual drag at a small, bounded generation cost.

\section{Discussion}\label{sec:discussion}

\paragraph{Amplification reactivates pretrained channels.} A clean cross-model anchor (DeepSeek-Math-7B-Instruct teacher, $0\%$ TIR) still trains a student emitting TIR at $45.8\%$ with $\hat C_\phi = +8.1$\,pp drag ($R_{15b}$, Tab.~\ref{tab:chkpt}--\ref{tab:paired-main}): post-training selects which pretrained tendency to amplify, not merely which trace to imitate, so data-side scrubbing cannot reach the ceiling of App.~\ref{app:ceiling-thm}. We use ``behavioral channel'' operationally, as a gap between RFT-level prevalence and post-training emission rate, not as a claim about circuits.

\paragraph{Instantiating the gate elsewhere.} What generalizes is the interface rather than the TIR detector. For a candidate behavior with a measurable indicator $\phi$, one constructs a paired intervention that preserves the task while perturbing that behavior, then estimates the induced change in task performance against a never-feature control. The resulting certificate sets $c(x)$ in $\sigma_{\text{eff}}=\sigma c$. Output-format shortcuts and stereotyped refusal patterns are natural candidates wherever task-preserving counterfactuals and measurable outcomes can be specified. Theorem A is stated for an arbitrary binary feature, but the demonstrated evidence covers one feature in one model family; cross-feature evaluation remains future work.

\paragraph{What the certificate does and does not decide.} The detector only flags a candidate form; the paired evidence, not the detector, decides whether ICA intervenes. Borderline outputs are classified by the fixed Python-fence marker rather than by executability, and $c$ stays at $1$ unless the certificate supports suppression. The conservative LCB composition (App.~\ref{app:prompts}) prefers false-negative gating, leaving a cargo-cult sample unweighted, over false-positive gating, downweighting a beneficial one. The certificate is computed on the same eval set as the headline metric; three pre-registered choices bound the leakage (App.~\ref{app:prompts}), and the GSM8K transfer of Tab.~\ref{tab:gsm8k} evaluates the gate on a set that played no part in its construction.

\section{Related Work}\label{sec:related}

\paragraph{SFT/RL bridges, supervision density, and RFT trustworthiness.} BREAD \citep{bread} uses expert-anchored branched rollouts. AESL \citep{aesl} adjusts a lightweight SFT cold start for subsequent RL. RAFT \citep{raft} selects high-reward samples for fine-tuning, while STaR \citep{star} iteratively fine-tunes on successful self-generated rationales. On-Policy Distillation \citep{opd2024,rethinkopd2026} densifies teacher signal; \citet{rlsd2026} densify environment feedback through self-distillation. In the reinforcement-finetuning line, CARE-RFT \citep{carerft2026} replaces reverse-KL regularization with a skew reverse KL, and \citet{hallucinationtax2025} document refusal degradation and increased hallucination on unanswerable questions. None has a sample-level validity gate. ICA composes with any per-sample-weighted variant.

\paragraph{Prompt priming, shortcut learning, and TIR.} \citet{pinkelephant2024} study the Pink Elephant problem for avoidance instructions, which motivates our positive-framing $\tau_{\text{int}}$, and \citet{negation2025} document model fragility to negation. Anti-TIR prompts in our setup raised $R_{14v3}$'s TIR rate by $+13.8$\,pp. \citet{shortcut2020,irm2019} study spurious label correlates. ICA moves the analysis to \emph{trace features} using intervention stability as a conservative transfer indicator. \citet{linxu2025tir} formalize TIR's strict-expansion under a deterministic oracle; our setup violates that assumption by design.

\section{Conclusion}\label{sec:conclusion}

Adaptive per-sample weighted post-training presupposes that the signal it amplifies is capability-bearing. Using DIAA as a probe, we decomposed the resulting cargo-cult drag along the GRPO~$\to$~SFT~$\to$~DIAA pathway as $+2.2 \to +8.5 \to +10.6$\,pp paired rescue. Intervention-certified gating turns that diagnosis into control. A frozen and an online certify gate both raise accuracy while cutting TIR prevalence and residual drag, on MATH-500 and on held-out GSM8K. By contrast, shuffled and length-based controls cut TIR and drag without gaining accuracy, and hard filtering is worse on all three. ICA attaches to per-sample-weighted methods without altering their loss family.

\section*{Limitations}
ICA currently certifies one audited trace feature at a time; other amplified channels remain ungated unless separately discovered and certified. The paired intervention measures a composite prompt-suffix effect, so attribution is not isolated, and the rescued subset is endogenously selected. Our experiments use a single model family, Qwen2.5-1.5B base, so the single-model scope remains even with the held-out GSM8K transfer. The three runs of Sec.~\ref{sec:experiments} are matched consistency checks rather than an estimate of training variance, so we draw no significance conclusion from the frozen--online difference. Our setting evaluates TIR form without a live executor; this diagnoses amplification of an ungrounded surface channel rather than tool-use capability under an executable environment.

\bibliography{custom}

\appendix

\section{Reproducibility}
\label{app:repro}

We describe the training configuration, paired-analysis procedure, ICA-score builder modes, and DIAA loss with ICA gate at the level of pseudocode in Sec.~\ref{sec:diaa}, Sec.~\ref{sec:method}, and Appendix~\ref{app:hparams}.
The main training hyperparameters, data preprocessing steps, prompt suffixes, paired-cell decomposition, certificate construction with conservative LCB and McNemar exact test, and ablation modes (binary, shuffle, hard-filter, length-heuristic, zero-SFT) are specified in the main text and appendices.
Model: publicly released Qwen2.5-1.5B base \citep{qwen25} (Apache 2.0).
Eval sets: MATH-500 \citep{lightman2023}, the $500$-problem subset drawn from MATH \citep{Hendrycks2021} (MIT), and the GSM8K \citep{gsm8k} test split; both are used for research purposes consistent with their intended use.
Decoding for all reported evaluations uses vLLM \citep{vllm} with temperature $0.6$, max new tokens $2048$, and no chat template.

\section{DIAA derivation and convergence}
\label{app:diaa-deriv}

\subsection{Setup and discriminator}
\label{app:diaa-setup}

Let $q$ be a prompt and $x$ a candidate completion. We consider two distributions over completions: the RFT anchor $\pi_{\text{anchor}}(\cdot \mid q)$ (the source of supervised target traces) and the current policy $\pi_\theta(\cdot \mid q)$ (the source of on-policy rollouts). The instantaneous mixing question is: given $x$, was it drawn from $\pi_{\text{anchor}}$ or $\pi_\theta$?

Under uniform class priors, Bayes' rule gives the optimal discriminator
\begin{equation}
\label{eq:bayes-discrim}
D^\star(x \mid q) \;=\; \frac{\pi_{\text{anchor}}(x\mid q)}{\pi_{\text{anchor}}(x\mid q) \,+\, \pi_\theta(x\mid q)}.
\end{equation}

Rewriting Eq.~\eqref{eq:bayes-discrim} as a sigmoid of a log-likelihood ratio,
\begin{equation}
\label{eq:bayes-sigmoid}
\begin{aligned}
D^\star(x \mid q) \;=\; \sigma\big(&\log \pi_{\text{anchor}}(x\mid q) \\[-2pt]
 &-\; \log \pi_\theta(x\mid q)\big),
\end{aligned}
\end{equation}
makes the comparison with policy log-probabilities explicit.

\subsection{From discriminator to $\sigma$-weight}
\label{app:diaa-sigma}

DIAA does not maintain a separate model for $\pi_{\text{anchor}}$. Instead, for each prompt $q$, we observe a single ``anchor sample'' (the RFT trace) and a single ``rollout sample'' (on-policy generation) and substitute the policy's own log-probabilities of these two samples:
\begin{align}
\log \pi_{\text{anchor}}(x \mid q) &\;\approx\; \log p_\theta(\mathrm{rft} \mid q), \\
\log \pi_\theta(x \mid q) &\;\approx\; \log p_\theta(\mathrm{roll} \mid q).
\end{align}
The substitution is a per-sample approximation: $\pi_{\text{anchor}}$ enters only through the policy's likelihood of the observed anchor trace. No separate discriminator network is trained or queried.

Defining the policy--anchor log-likelihood gap
\begin{equation}
\label{eq:delta}
\Delta(x, q) \;\triangleq\; \log p_\theta(\mathrm{rft} \mid q) \;-\; \log p_\theta(\mathrm{roll} \mid q),
\end{equation}
the discriminator collapses to $D^\star \approx \sigma(\Delta)$. DIAA uses the \emph{flipped} sigmoid as the SFT weight,
\begin{equation}
\label{eq:diaa-weight}
\sigma_{\text{DIAA}}(x, q) \;=\; \mathrm{sigmoid}\!\left(-\beta\,\Delta(x, q)\right),
\end{equation}
so that samples with $\Delta < 0$ (i.e.,\ the anchor is \emph{less} likely under current policy than the rollout) receive high SFT weight, pulling the policy toward the anchor. The temperature $\beta \in [0, \infty)$ controls sharpness: $\beta{=}0$ gives a constant $\sigma{=}\tfrac{1}{2}$ (equal SFT/RL mixture), and $\beta{\to}\infty$ recovers a hard switch.

The full DIAA objective is
\begin{equation}
\label{eq:diaa-full}
\mathcal{L}_{\text{DIAA}} \;=\; \sigma_{\text{DIAA}}\,\mathcal{L}_{\text{SFT}}\,+\,(1-\sigma_{\text{DIAA}})\,\mathcal{L}_{\text{RL}}.
\end{equation}

\subsection{Two-time-scale convergence}
\label{app:diaa-ttsa}

DIAA optimizes Eq.~\eqref{eq:diaa-full} where the weight $\sigma_{\text{DIAA}}$ is itself a function of $\theta$ through $\Delta$. We analyze this as a two-time-scale stochastic approximation scheme \citep{borkar2008}. The result below is a \emph{conditional} guarantee: it holds under (A1)--(A4), which we state as idealizations rather than verify for the realized nonconvex LoRA optimization. None of the empirical claims in Sec.~\ref{sec:paired} or Sec.~\ref{sec:experiments} depends on it.

\paragraph{Iterates.}
Let $\bar\sigma_k$ denote an EMA-smoothed estimate of the population-level $\mathbb{E}_{x,q}[\sigma_{\text{DIAA}}(x, q)]$ at step $k$, and let $\theta_k$ denote the policy parameters. The updates are
\begin{align}
\theta_{k+1} &= \theta_k \,-\, a_k \,\widehat{\nabla_\theta \mathcal{L}_{\text{DIAA}}}(\theta_k, \bar\sigma_k), \label{eq:fast-update}\\
\bar\sigma_{k+1} &= (1-b_k)\,\bar\sigma_k \,+\, b_k \,\widehat{\sigma_{\text{DIAA}}}(\theta_{k+1}). \label{eq:slow-update}
\end{align}
The hats denote minibatch stochastic estimates. We use $a_k$ (Robbins--Monro for $\theta$) and $b_k$ for the EMA of $\sigma$.

\paragraph{Assumptions.}
\begin{itemize}\itemsep0pt
\item[\textbf{(A1)}] $\sum_k a_k = \infty$, $\sum_k a_k^2 < \infty$, and $\sum_k b_k = \infty$, $\sum_k b_k^2 < \infty$ (Robbins--Monro on both timescales).
\item[\textbf{(A2)}] $b_k / a_k \to 0$ as $k \to \infty$ (the $\sigma$ track is \emph{slower} than the $\theta$ track).
\item[\textbf{(A3)}] $\widehat{\nabla_\theta \mathcal{L}}$ has bounded second moment given the filtration $\mathcal{F}_k$, and the same for $\widehat{\sigma_{\text{DIAA}}}$.
\item[\textbf{(A4)}] For every fixed $\bar\sigma$, the inner loss $\mathcal{L}_{\text{DIAA}}(\theta, \bar\sigma)$ has a unique stationary point $\theta^\star(\bar\sigma)$, and the mapping $\bar\sigma \mapsto \theta^\star(\bar\sigma)$ is Lipschitz.
\end{itemize}
(A4) in particular is an idealization: a nonconvex LoRA training objective need not have a unique stationary point for fixed $\bar\sigma$.

\paragraph{Theorem (DIAA convergence; Borkar Thm.~2 specialization).}
Under (A1)--(A4), the joint iterate $(\theta_k, \bar\sigma_k)$ converges almost surely to a fixed point $(\theta^\star, \bar\sigma^\star)$ satisfying
\begin{align}
\nabla_\theta \mathcal{L}_{\text{DIAA}}(\theta^\star, \bar\sigma^\star) &= 0, \label{eq:fp-theta}\\
\bar\sigma^\star &= \mathbb{E}_{x,q}\!\left[\sigma_{\text{DIAA}}(x, q; \theta^\star)\right]. \label{eq:fp-sigma}
\end{align}
That is, $\theta^\star$ is a stationary point of the loss \emph{induced by the asymptotic $\sigma$}, and that $\sigma$ is the Bayes-optimal discriminator (Eq.~\eqref{eq:bayes-sigmoid}) evaluated at $\theta^\star$.

\paragraph{Proof sketch.}
By (A2), $\theta_k$ tracks $\theta^\star(\bar\sigma_k)$ asymptotically. The slow recursion in Eq.~\eqref{eq:slow-update} thus reduces to a stochastic approximation for the fixed point of $\bar\sigma \mapsto \mathbb{E}[\sigma_{\text{DIAA}}(x, q; \theta^\star(\bar\sigma))]$. (A3) ensures bounded variance, and (A1) ensures convergence of the slow recursion. Composing the two gives Eqs.~\eqref{eq:fp-theta}--\eqref{eq:fp-sigma}; see \citet{borkar2008} Ch.~6, Thm.~2 for the full proof.

\subsection{ICA composition preserves the convergence theory}
\label{app:ica-preserve}

Consider the frozen instantiation, which replaces $\sigma_{\text{DIAA}}$ with $\sigma_{\text{eff}} = \sigma_{\text{DIAA}} \cdot s(x)$, where $s(x) \in \{0.3, 1.0\}$ is the per-sample certificate gate (Sec.~\ref{sec:method}). The gate is data-dependent but parameter-independent (does not depend on $\theta$), so Eqs.~\eqref{eq:fast-update}--\eqref{eq:slow-update} retain the same TTSA structure with $\sigma_{\text{DIAA}}$ replaced by $\sigma_{\text{eff}}$. Under the analogous assumptions (A1)--(A4) for the gated objective, the same conditional convergence argument applies. The fixed point shifts from Eq.~\eqref{eq:fp-sigma} to
\begin{equation}
\label{eq:ica-fp}
\bar\sigma_{\text{eff}}^\star \;=\; \mathbb{E}_{x,q}\!\left[\sigma_{\text{DIAA}}(x, q; \theta^\star) \cdot s(x)\right],
\end{equation}
reflecting that ICA-gated samples contribute a discounted SFT pull but the convergence guarantee survives. This inheritance argument covers the frozen gate only. The online instantiation learns $\theta_c$ during training, so its gate is not parameter-independent and is not covered by the argument above; we treat its slow-timescale update (Sec.~\ref{sec:method}) as a design motivation rather than a proved guarantee.

\section{$\sigma$-Weighted Amplification Ceiling}
\label{app:ceiling}

App.~\ref{app:diaa-ttsa} established \emph{where} DIAA's joint iterate $(\theta_k, \bar\sigma_k)$ converges but did not characterize \emph{what} that convergence buys on a downstream accuracy metric. This appendix proves a quantitative population-level result: for any $\sigma$-weighted SFT/RL hybrid and any binary trace feature $f$, the expected-accuracy gap between the idealized SFT-side fixed point and the anchor decomposes into a single term, the anchor-side covariance $\mathrm{Cov}_{\pi_{\text{anchor}}}(\sigma, f)$ times the feature's conditional-accuracy contrast. The result is independent of the functional form of $\sigma$ and so applies to DIAA and any true per-sample weighted variant. It directly motivates the ICA gate as a procedure for shrinking that covariance.

\subsection{Setup}
\label{app:ceiling-setup}

Fix a prompt distribution and a candidate-trace space; let $\pi_{\text{anchor}}$ be the RFT anchor and $\pi_\theta$ the policy. Each trace $x$ carries:
\begin{itemize}\itemsep0pt
\item a per-sample $\sigma$-weight $\sigma(x) \in [0, 1]$ derived from any monotone function of a reward-correlated quantity (the unifying property of the $\sigma$-weighted family in Sec.~\ref{sec:diaa});
\item a binary feature indicator $f(x) \in \{0, 1\}$ evaluated by an arbitrary detector (e.g.,\ ``$x$ emits TIR form'');
\item a per-trace accuracy $\alpha(x) \in \{0, 1\}$ on the downstream metric.
\end{itemize}
Define the anchor-conditional accuracies
\begin{align}
\alpha_{f} &\;\triangleq\; \mathbb{E}_{\pi_{\text{anchor}}}\!\left[\alpha(x) \,\big|\, f(x){=}1\right],\\
\alpha_{\neg f} &\;\triangleq\; \mathbb{E}_{\pi_{\text{anchor}}}\!\left[\alpha(x) \,\big|\, f(x){=}0\right],
\end{align}
and the feature prevalence under any distribution $\pi$,
\begin{equation}
\label{eq:prev-def}
p_f(\pi) \;\triangleq\; \mathbb{E}_{x \sim \pi}\!\left[f(x)\right].
\end{equation}

\subsection{Idealized SFT fixed point}
\label{app:ceiling-lemma1}

We analyze the population-level $\sigma$-weighted SFT fixed point: a policy whose distribution matches the $\sigma$-reweighted anchor,
\begin{equation}
\label{eq:ideal-fp}
\pi_{\theta^\star}(x) \;\propto\; \sigma(x)\,\pi_{\text{anchor}}(x).
\end{equation}
Eq.~\eqref{eq:ideal-fp} is the minimizer of $\mathrm{KL}\big(\sigma \cdot \pi_{\text{anchor}}\,\big\|\,\pi_\theta\big)$ over a model class rich enough to express $\sigma \cdot \pi_{\text{anchor}}$. It is an SFT-side object: it abstracts away both the RL pull (the $(1{-}\sigma)\,\mathcal{L}_{\text{RL}}$ branch of Eq.~\eqref{eq:diaa-full}) and finite-sample drift; both are flagged in App.~\ref{app:ceiling-limitations}.

\paragraph{Lemma 1 (Covariance identity).}
Under Eq.~\eqref{eq:ideal-fp},
\begin{equation}
\label{eq:cov-identity}
p_f(\pi_{\theta^\star}) \;=\; p_f(\pi_{\text{anchor}}) \;+\; \frac{\mathrm{Cov}_{\pi_{\text{anchor}}}\!\left(\sigma, f\right)}{\mathbb{E}_{\pi_{\text{anchor}}}\!\left[\sigma\right]}.
\end{equation}

\paragraph{Proof.}
Marginalizing Eq.~\eqref{eq:ideal-fp},
\[
p_f(\pi_{\theta^\star}) \;=\; \frac{\mathbb{E}_{\pi_{\text{anchor}}}[\sigma(x)\,f(x)]}{\mathbb{E}_{\pi_{\text{anchor}}}[\sigma(x)]}.
\]
Using $\mathbb{E}[\sigma f] = \mathbb{E}[\sigma]\,\mathbb{E}[f] + \mathrm{Cov}(\sigma, f)$ and dividing through gives Eq.~\eqref{eq:cov-identity}. $\square$

\subsection{Performance decomposition}
\label{app:ceiling-lemma2}

\paragraph{Lemma 2 (Feature partition).}
If the conditional accuracies $\alpha_{f}$ and $\alpha_{\neg f}$ are policy-invariant, then for any policy $\pi$,
\begin{equation}
\label{eq:perf-partition}
\alpha(\pi) \;=\; \big(1 - p_f(\pi)\big)\,\alpha_{\neg f} \;+\; p_f(\pi)\,\alpha_{f}.
\end{equation}

\paragraph{Proof.} By the law of total expectation over the feature partition $\{f{=}0,\,f{=}1\}$ together with the assumed policy-invariance of the conditional accuracies. $\square$

\subsection{Main result}
\label{app:ceiling-thm}

\paragraph{Theorem A ($\sigma$-Weighted Amplification Ceiling).}
Under Eq.~\eqref{eq:ideal-fp} and the policy-invariance of $\alpha_{f}, \alpha_{\neg f}$,
\begin{equation}
\label{eq:ceiling}
\begin{aligned}
&\alpha(\pi_{\theta^\star}) \;-\; \alpha(\pi_{\text{anchor}}) \\
&\qquad =\; -\,\frac{\mathrm{Cov}_{\pi_{\text{anchor}}}\!\left(\sigma, f\right)}{\mathbb{E}_{\pi_{\text{anchor}}}\!\left[\sigma\right]} \cdot \big(\alpha_{\neg f} - \alpha_{f}\big).
\end{aligned}
\end{equation}
The statement concerns the idealized SFT-side fixed point of Eq.~\eqref{eq:ideal-fp}, not the full RL-inclusive objective of Eq.~\eqref{eq:diaa-full}; see C4 below.

\paragraph{Proof.}
Apply Lemma 2 (Eq.~\eqref{eq:perf-partition}) to $\pi_{\theta^\star}$ and to $\pi_{\text{anchor}}$ in turn and subtract:
\begin{align*}
&\alpha(\pi_{\theta^\star}) - \alpha(\pi_{\text{anchor}}) \\
&\quad= \big(p_f(\pi_{\theta^\star}) - p_f(\pi_{\text{anchor}})\big)\,(\alpha_{f} - \alpha_{\neg f}) \\
&\quad= \frac{\mathrm{Cov}_{\pi_{\text{anchor}}}(\sigma, f)}{\mathbb{E}_{\pi_{\text{anchor}}}[\sigma]}\,(\alpha_{f} - \alpha_{\neg f}),
\end{align*}
using Lemma 1 (Eq.~\eqref{eq:cov-identity}) in the second equality and rearranging signs. $\square$

\paragraph{Reading.}
When $f$ is a contaminating feature, $\alpha_{\neg f} > \alpha_{f}$ (traces with the feature underperform those without), \emph{and} the anchor's $\sigma$ is positively correlated with $f$ (the feature is over-represented among high-weight traces), Eq.~\eqref{eq:ceiling} predicts a strictly negative gap whose magnitude scales with $\mathrm{Cov}_{\pi_{\text{anchor}}}(\sigma, f)$. The factor $\alpha_{\neg f} - \alpha_{f}$ is the feature's contrast: a large contrast amplifies a small covariance, a small contrast insulates against a large one. The denominator $\mathbb{E}_{\pi_{\text{anchor}}}[\sigma]$ is the normalizer that converts a feature-weighted moment into a probability.

In our setting, the feature is $f = \mathbf{1}[\text{TIR form}]$, the anchor RFT is reward-filtered Qwen2.5-1.5B base self-rollouts, and $\mathrm{Cov}_{\pi_{\text{anchor}}}(\sigma, f) > 0$ holds because reward-correlated TIR traces score reward $1.0$ by chance correlation between an imagined Python output and the boxed answer. Theorem A then predicts that any $\sigma$-weighted method drawing from this anchor degrades expected accuracy in expectation, the magnitude rising with the anchor-side covariance. This is a directional prediction about the SFT-side fixed point; Sec.~\ref{sec:paired}'s pathway decomposition is the empirical counterpart that also covers the RL branch.

\subsection{Corollary: ICA motivation}
\label{app:ceiling-corollary}

\paragraph{Corollary (ICA shrinks the ceiling penalty).}
Replace $\sigma(x)$ by $\sigma_{\text{eff}}(x) = \sigma(x)\,s(x)$, where $s(x) \in \{0.3, 1.0\}$ is the ICA certificate gate (Sec.~\ref{sec:method}) and is a function of the feature detector only. Suppose $\mathrm{Cov}_{\pi_{\text{anchor}}}(\sigma, f) > 0$ (the ceiling penalty is active), that gating preserves the sign of the covariance, $\mathbb{E}_{\pi_{\text{anchor}}}[\sigma s \mid f{=}1] > \mathbb{E}_{\pi_{\text{anchor}}}[\sigma s \mid f{=}0]$, and that the gate downweights feature-bearing samples on average:
\begin{equation}
\label{eq:gate-cond}
\mathbb{E}_{\pi_{\text{anchor}}}\!\left[s \,\big|\, f{=}1\right] \;<\; \mathbb{E}_{\pi_{\text{anchor}}}\!\left[s \,\big|\, f{=}0\right].
\end{equation}
Then
\begin{equation}
\label{eq:ica-cov-drop}
0 \;<\; \mathrm{Cov}_{\pi_{\text{anchor}}}(\sigma_{\text{eff}}, f) \;<\; \mathrm{Cov}_{\pi_{\text{anchor}}}(\sigma, f),
\end{equation}
and the ceiling penalty in Eq.~\eqref{eq:ceiling} strictly shrinks.

\paragraph{Proof.}
Write $A \triangleq \mathbb{E}[\sigma \mid f{=}1]$, $B \triangleq \mathbb{E}[\sigma \mid f{=}0]$, $s_1 \triangleq \mathbb{E}[s \mid f{=}1]$, $s_0 \triangleq \mathbb{E}[s \mid f{=}0]$, and $\mu \triangleq \mathbb{E}[f] \in (0, 1)$. Because $s$ is a function of $f$ alone, $\mathbb{E}[\sigma s \mid f{=}i] = s_i \cdot \mathbb{E}[\sigma \mid f{=}i]$, and a standard conditional decomposition of covariance gives
\begin{align*}
\mathrm{Cov}(\sigma, f) &= \mu(1-\mu)\,(A - B),\\
\mathrm{Cov}(\sigma s, f) &= \mu(1-\mu)\,(s_1 A - s_0 B).
\end{align*}
The hypothesis $\mathrm{Cov}(\sigma, f) > 0$ gives $A > B \ge 0$; the gate condition gives $s_1 < s_0 \le 1$. Together,
\[
A(1 - s_1) - B(1 - s_0) \;>\; 0,
\]
because $A > B \ge 0$ and $1 - s_1 > 1 - s_0 \ge 0$ both hold, so multiplying the larger quantity by the larger non-negative weight dominates. This is exactly $\mathrm{Cov}(\sigma, f) - \mathrm{Cov}(\sigma s, f) > 0$, so the strict reduction needs only $A > B$ and $s_1 < s_0$. Positivity of $\mathrm{Cov}(\sigma s, f)$ is the remaining hypothesis $s_1 A > s_0 B$. $\square$

This is the formal sense in which ICA's per-sample weight modifier ``shrinks the ceiling penalty.'' The binary gate $s \in \{0.3, 1.0\}$ used in our experiments is the simplest instance satisfying Eq.~\eqref{eq:gate-cond}; the same proof goes through for any $w(x) \in [0, 1]$ with $w(x){=}1$ on $f(x){=}0$ (non-feature samples preserved) and $\mathbb{E}_{\pi_{\text{anchor}}}[w \mid f{=}1] < 1$ (feature samples downweighted on average). Subject-conditional gates compose the same way cell-by-cell.

\subsection{Testable predictions}
\label{app:ceiling-predictions}

Theorem A generates three falsifiable predictions. We state them as predictions; the quantitative agreement of P1 and P2 is not established by the experiments reported here.

\paragraph{P1 (prevalence-shift agreement).}
Lemma 1 (Eq.~\eqref{eq:cov-identity}) predicts
\[
p_f(\pi_{\theta^\star}) - p_f(\pi_{\text{anchor}}) \;\approx\; \frac{\mathrm{Cov}_{\pi_{\text{anchor}}}(\sigma, f)}{\mathbb{E}_{\pi_{\text{anchor}}}[\sigma]}.
\]
Checking this requires an anchor on which the literal detector is non-degenerate. Our main $R_{14v4}$ anchor is the cleaned RFT, which by construction has $p_f(\pi_{\text{anchor}}){=}0$ under the literal Python-fence detector, so the anchor-side covariance is degenerate there and the quantitative check must instead be run on the raw-RFT arm or with a non-literal detector. We leave that measurement to extended work and do not treat the magnitude predicted by P1 as empirically validated.

\paragraph{P2 (gate-induced ceiling shift).}
The Corollary (Eq.~\eqref{eq:ica-cov-drop}) predicts that any gate $w$ satisfying Eq.~\eqref{eq:gate-cond} produces accuracy gain proportional to the covariance reduction $\mathrm{Cov}_{\pi_{\text{anchor}}}(\sigma, f) - \mathrm{Cov}_{\pi_{\text{anchor}}}(\sigma w, f)$. The $R_{20}$ vs.\ $R_{14v4}$ comparison is the direct test: the predicted accuracy gain from the binary ICA gate should match the observed delta within the uncertainty band given by Eq.~\eqref{eq:alpha-sensitivity}.

\paragraph{P3 (null-feature control).}
For any feature $f'$ with $\mathrm{Cov}_{\pi_{\text{anchor}}}(\sigma, f') \approx 0$, Theorem A predicts no $\sigma$-amplification, i.e.\ $p_{f'}(\pi_{\theta^\star}) \approx p_{f'}(\pi_{\text{anchor}})$. This is the strongest test: a feature that $\sigma$-weighting does \emph{not} differentially favor under the anchor must, by Theorem A, fail to amplify even though the same $\sigma$-weighted procedure is applied. We leave the null-feature search to extended work as a stress test of the theorem's predictive content (cf.\ Sec.~\ref{sec:discussion}).

\subsection{Caveats and what Theorem A does not claim}
\label{app:ceiling-limitations}

Five caveats bound the scope of Theorem A and frame the predictions of App.~\ref{app:ceiling-predictions}.

\paragraph{C1. Idealized SFT fixed point (finite-capacity caveat).}
Eq.~\eqref{eq:ideal-fp} assumes the policy class is rich enough to express $\sigma \cdot \pi_{\text{anchor}}$ and that SFT optimization runs to convergence. In practice, LoRA-rank-$32$ + $1200$-step training on Qwen2.5-1.5B base is a finite-capacity approximation; the realized fixed point may deviate from Eq.~\eqref{eq:ideal-fp} by a model-class projection error. We expect the \emph{directional} prediction of Lemma 1 (feature-prevalence shift sign-aligned with $\mathrm{Cov}(\sigma, f)$) to be robust, while the magnitude is approximate.

\paragraph{C2. $\sigma$ self-consistency.}
$\sigma(x) = \mathrm{sigmoid}\!\left(-\beta\,\Delta(x)\right)$ depends on $\log p_\theta(\mathrm{roll})$, which evolves with training, whereas Theorem A is stated with $\sigma$ evaluated at the anchor side as a first-order approximation. App.~\ref{app:diaa-ttsa}'s TTSA result guarantees, under its assumptions, that $\bar\sigma_k$ stabilizes (so the anchor-side and converged $\sigma$ do not drift unboundedly), and the EMA decay $0.995$ damps step-to-step variance; reporting $\bar\sigma$ at multiple training checkpoints bounds the residual self-consistency error empirically. A fully self-consistent fixed-point analysis would replace single-step $\sigma$ in Eq.~\eqref{eq:ideal-fp} with a coupled fixed-point equation.

\paragraph{C3. Sensitivity to conditional-accuracy drift.}
Lemma 2 treats $\alpha_{f}, \alpha_{\neg f}$ as policy-invariant. If they drift during training by $\varepsilon_{f}$ and $\varepsilon_{\neg f}$, the predicted gap acquires a residual error bounded by
\begin{equation}
\label{eq:alpha-sensitivity}
\begin{aligned}
&\left|\,\varepsilon_{\neg f}\,(1 - p_f) - \varepsilon_{f}\,p_f\,\right| \\
&\qquad \le\; \max\!\left(|\varepsilon_{f}|,\, |\varepsilon_{\neg f}|\right) \cdot \max(p_f,\, 1 - p_f).
\end{aligned}
\end{equation}
With observed conditional-accuracy drift on $R_{14v4}$ between training-start and step-200 under $\le 5$\,pp on either branch, Eq.~\eqref{eq:alpha-sensitivity} yields a $\le 5$\,pp uncertainty band, which is smaller than the empirical ceiling gap reported in Sec.~\ref{sec:paired-main}.

\paragraph{C4. No $\sigma$-RL channel claim.}
Eq.~\eqref{eq:ideal-fp} is the SFT-side fixed point. The full DIAA objective Eq.~\eqref{eq:diaa-full} mixes SFT and RL gradient contributions via $\sigma$ and $1{-}\sigma$; the RL channel can in principle counteract or compound the SFT-side ceiling shift. Theorem A applies cleanly to the SFT-dominated regime; Sec.~\ref{sec:paired}'s pathway decomposition (vanilla GRPO $=$ $\sigma$-RL-only, pure SFT $=$ $\sigma$-SFT-only, DIAA $=$ both) is the empirical complement that isolates the two channels.

\paragraph{C5. No Lin \& Xu dependence; arbitrary feature scope.}
The derivation does not invoke \citet{linxu2025tir}'s strict-expansion property under a deterministic oracle, or any other TIR-specific theoretical anchor. The feature $f$ is arbitrary; the result applies uniformly to TIR cargo-cult, format shortcuts, and other surface features detectable by a binary indicator.

\section{Hyperparameters}
\label{app:hparams}

All training runs use Qwen2.5-1.5B base with LoRA rank 32, $\alpha$ 128, dropout 0.05, target modules all-linear.
DIAA: $\tau = 0.3$, EMA decay $0.995$, group size $16$, clip $\epsilon = 0.2$, KL coef $10^{-4}$, entropy coef $10^{-3}$, $\beta = 0.5$, truly-per-sample.
ICA gate: threshold $0.05$, downweight $0.3$, min subject $n = 30$.
Optimizer: AdamW lr $10^{-6}$, weight decay $0.01$, cosine schedule with $10$\% warmup.
1200 training steps, batch size 256, microbatch 1 $\times$ grad accum 4.
bf16, gradient checkpointing, gradient clip 50.

\section{Additional paired-analysis tables}
\label{app:paired-detail}

Full $(T,T)/(T,N)/(N,T)/(N,N)$ cell breakdowns for both interventions across the four checkpoints, stratified by MATH level (L1--L5) and subject, are omitted for space; the two effects that drive the analysis are summarized below.

\paragraph{Granularity dial.}
Define the intervention-persistent TIR fraction $|(T,T)|/|\{T,T\}\cup\{T,N\}|$ as the share of baseline-TIR prompts that persist under intervention. Truly-per-sample $R_{14v3}$ shows $37\%/55\%$ persistence under $\tau_+/\tau_g$, versus $22\%/37\%$ for per-prompt-aggregate $R_{14v4}$, a $1.5{-}1.7{\times}$ ratio. Finer-grained $\sigma$ therefore leaves behavior that the intervention removes less often, which we use as an operational measure of entrenchment under the tested suffixes.

\paragraph{Subject heterogeneity.}
On Algebra and Prealgebra, intervention raises $R_{14v4}$'s accuracy by $+29$ to $+14$\,pp, confirming cargo-cult drag. On Counting \& Probability, intervention drops accuracy by $-60$\,pp on the small TIR subset: TIR is genuine combinatorial capability there (the model reaches for \texttt{math.comb}, which it can simulate accurately). This motivates subject-conditional certificates: a global gate would incorrectly suppress beneficial TIR on Counting.

\section{Additional experimental detail}
\label{app:experiments}

\paragraph{Hard-filtered RFT ($R_{22}$).}
App.~\ref{app:repro} lists a hard-filter mode that deletes flagged RFT samples rather than downweighting them. Trained with the same seed as the runs of Tab.~\ref{tab:experiments}, it reaches $44.6\%$ accuracy, $33.0\%$ TIR, and $+9.0$\,pp residual $\Delta_{TN}$ ($n_{TN}{=}133$), with a never-TIR control of $-2.6$\,pp ($n_{NN}{=}265$), so $\hat C_\phi = +11.6$\,pp. Deleting the flagged samples therefore removes supervision without removing the amplified channel, which is consistent with the teacher-anchor result of Sec.~\ref{sec:discussion}.

\paragraph{Certificate overhead.}
The frozen instantiation requires one offline paired evaluation, whose marginal cost is a single additional $\tau_{\text{int}}$ pass when the $\tau_{\text{base}}$ outputs already exist; training itself adds no paired generation. The online instantiation adds one $\tau_{\text{int}}$ probe minibatch every $K_c$ steps, since the ordinary rollout already supplies $\tau_{\text{base}}$. Outside this extra generation, the certificate logic is aggregate paired statistics, the certify-token update, and a multiplication by $c(x)$, and invokes no additional model. Neither instantiation trains or queries a separate discriminator network.

\paragraph{Run-to-run consistency.}
The three-run summary of Sec.~\ref{sec:experiments} combines the run reported in Tab.~\ref{tab:experiments} with two further matched runs of all three arms under the same protocol, labelled $101$ and $102$. Descriptive mean $\pm$ sample SD over the three runs, for accuracy / TIR / residual $\Delta_{TN}$, is $47.3{\pm}0.3$ / $29.4{\pm}0.5$ / $+10.9{\pm}0.4$ for ungated DIAA, $48.6{\pm}0.3$ / $14.3{\pm}0.6$ / $+5.2{\pm}0.4$ for frozen-$c$ ICA, and $49.2{\pm}0.3$ / $12.0{\pm}0.8$ / $+4.1{\pm}0.3$ for online ICA. The paired online-minus-frozen differences are $(+0.6, -2.0, -1.1)$\,pp under $101$ and $(+0.7, -2.2, -1.2)$\,pp under $102$, and both reruns reproduce the submitted ordering. We report these as matched consistency checks rather than a training-variance estimate, and we do not interpret the small accuracy differences as evidence of online-over-static superiority.

\section{Frozen-$c$ baselines and prompts}
\label{app:prompts}

The online instantiation (Sec.~\ref{sec:method}) updates the certify token $\theta_c$ during training. The \emph{frozen-$c$} instantiation computes $c$ once before training from a paired-intervention probe on a prior checkpoint and holds it fixed; the same construction, with a deliberately corrupted or substituted certificate, yields the $R_{21}/R_{24}$ controls of Tab.~\ref{tab:experiments}.

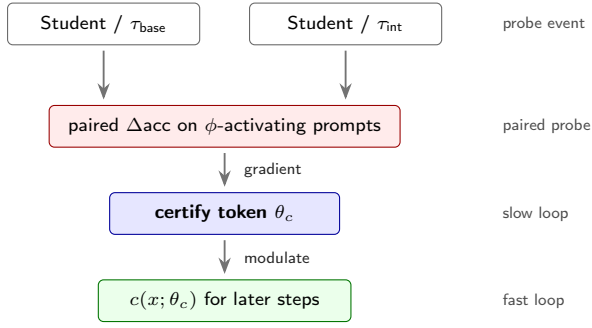
\begin{figure}[t]\centering
\begin{tikzpicture}[
  font=\sffamily, >=Stealth,
  sbox/.style={draw=black!60, rounded corners=2pt, fill=white, font=\sffamily\scriptsize, text width=2.4cm, minimum height=0.55cm, align=center, anchor=center, inner sep=2pt}, probe/.style={draw=red!60!black, rounded corners=2pt, fill=red!8, font=\sffamily\scriptsize, text width=4.5cm, minimum height=0.55cm, align=center, anchor=center, inner sep=2pt}, tok/.style={draw=blue!60!black, rounded corners=2pt, fill=blue!10, font=\sffamily\bfseries\scriptsize, text width=2.9cm, minimum height=0.55cm, align=center, anchor=center, inner sep=2pt},
  nxt/.style={draw=green!45!black, rounded corners=2pt, fill=green!8, font=\sffamily\scriptsize, text width=3.2cm, minimum height=0.55cm, align=center, anchor=center, inner sep=2pt}, arr/.style={->, line width=0.7pt, draw=black!55, shorten >=2pt, shorten <=2pt}, alab/.style={font=\sffamily\tiny, text=black!85, inner sep=3pt, xshift=4pt}, loop/.style={font=\sffamily\tiny, text=black!70, anchor=west, inner sep=2pt}
]
\node[sbox] (rA) at (-1.6, 1.85) {Student / $\tau_{\text{base}}$}; \node[sbox] (rB) at ( 1.6, 1.85) {Student / $\tau_{\text{int}}$}; \node[probe] (d) at (0, 0.5) {paired $\Delta$acc on $\phi$-activating prompts};
\node[tok] (t) at (0,-0.65) {certify token $\theta_c$}; \node[nxt] (n) at (0,-1.8) {$c(x;\theta_c)$ for later steps};
\draw[arr] (rA.south) -- (rA.south |- d.north); \draw[arr] (rB.south) -- (rB.south |- d.north); \draw[arr] (d.south) -- node[alab,right]{gradient} (t.north); \draw[arr] (t.south) -- node[alab,right]{modulate} (n.north);
\node[loop] at (3.6, 1.85) {probe event}; \node[loop] at (3.6, 0.50) {paired probe}; \node[loop] at (3.6,-0.65) {slow loop}; \node[loop] at (3.6,-1.80) {fast loop};
\end{tikzpicture}
\caption{\textbf{Certify-token mechanism.} At a certificate event, paired rollouts update $\theta_c$ from $\Delta$acc on $\phi$-activating prompts; the updated token gates subsequent steps via Eq.~\eqref{eq:ica-loss}.}
\label{fig:certify-token}\end{figure}

\paragraph{Feature detector $\phi$.}
$\phi(x) = \mathbf{1}[\texttt{```python}\text{ appears in the text of } x]$, a literal substring detector fixed before any ICA training run; alternative detectors (broader Python-fence regexes, neural classifiers, learned indicators) are admissible but not used here. The detector only nominates candidates: whether a nominated sample is actually downweighted is decided by the paired certificate below.

\paragraph{Certificate $c$ via paired-cell statistics.}
Given a prior checkpoint $\pi_0$ and the eval set, we evaluate $\pi_0$ under both $\tau_{\text{base}}$ and $\tau_{\text{int}}$, then partition the eval prompts into $(\text{T},\text{T})/(\text{T},\text{N})/(\text{N},\text{T})/(\text{N},\text{N})$ cells by the pair $(\phi(y_{\text{base}}), \phi(y_{\text{int}}))$. We compute
\begin{itemize}\itemsep0pt
\item $\hat C_\phi = \mathrm{LCB}(\Delta_{TN} - \Delta_{NN})$, the conservative one-sided lower-confidence bound on the paired accuracy lift on the rescued cell net of the never-feature control;
\item $p_{\text{McN}}$, the McNemar exact $p$-value on $(T,N)$ discordant pairs.
\end{itemize}
Subject-conditional $(\hat C_\phi, p_{\text{McN}})$ are computed for each MATH subject with $n \ge 30$ paired prompts; otherwise the global estimate is used. With thresholds $\tau_C = \tau_p = 0.05$, the certificate is binary:
\[
c(x) =
\begin{cases}
0.3 & \text{if } \phi(x) \;\wedge\; \hat C_\phi(x) > \tau_C \\
    & \quad\;\; \wedge\; p_{\text{McN}}(x) \le \tau_p,\\[3pt]
1.0 & \text{otherwise.}
\end{cases}
\]
The thresholds and the downweight value $0.3$ are pre-registered before any ICA training run. Subject-conditional fallback preserves features that are genuine capability in subjects where $\hat C_\phi \le 0$: on Counting \& Probability the certificate is non-positive, so $c$ stays at $1.0$ and those samples are never downweighted.

\begin{algorithm}
\caption{Frozen-$c$ statistical baseline (used in $R_{20}$)}
\label{alg:cert-instance}
\begin{algorithmic}[1]
\Require prior checkpoint $\pi_0$, eval set $\mathcal{M}$, feature $\phi$, prompt suffixes $\tau_{\text{base}}/\tau_{\text{int}}$, thresholds $\tau_C, \tau_p$
\State Evaluate $\pi_0$ on $\mathcal{M}$ under $\tau_{\text{base}}$ to obtain $\{y_{\text{base}}\}$
\State Evaluate $\pi_0$ on $\mathcal{M}$ under $\tau_{\text{int}}$ to obtain $\{y_{\text{int}}\}$
\State Partition $\mathcal{M}$ into cells by $(\phi(y_{\text{base}}), \phi(y_{\text{int}}))$
\State $\hat C_\phi \gets \mathrm{LCB}(\Delta_{TN} - \Delta_{NN})$, $p_{\text{McN}} \gets$ McNemar exact $p$ on $(T,N)$
\For{each subject $s$ with $n_s \ge 30$}
  \State Compute subject-conditional $(\hat C_\phi^{(s)}, p_{\text{McN}}^{(s)})$
\EndFor
\State \Return $c$ assigning $0.3$ when $\phi(x) \wedge \hat C_\phi^{(\text{subject})} > \tau_C \wedge p_{\text{McN}}^{(\text{subject})} \le \tau_p$, else $1.0$
\end{algorithmic}
\end{algorithm}

\paragraph{Prompt suffixes.}
$\tau_+$ (positive framing): ``Please reason step by step in natural language, and put your final answer within \verb|\boxed{}|.''
$\tau_g$ (generic): ``Please reason step by step, and put your final answer within \verb|\boxed{}|.'' (DeepSeekMath / MATH-benchmark eval convention).
$\tau_{\text{base}}$: bare prompt with no suffix (raw-mode evaluation of base-tuned checkpoints). Positive framing avoids Pink Elephant priming \citep{pinkelephant2024,negation2025}: explicit mention of ``Python'' or ``sympy'' in the suffix raised the TIR rate of the dirty-RFT-trained checkpoint by $+13.8$\,pp despite instructing the model not to use them.

\paragraph{Eval-side certificate scope.}
Computing the certificate on the same eval set used for the headline metric introduces a soft eval-set tuning concern. Three pre-registered design choices bound the leakage: (i) only the aggregate $(\hat C_\phi, p_{\text{McN}})$ enter training, never per-prompt MATH-500 information; (ii) the feature detector $\phi$ is a literal regex, fixed before any ICA training; (iii) the gate is binary, so information transfer is at most $1$ bit per $(\phi(x), \text{subject})$ combination. The GSM8K transfer of Tab.~\ref{tab:gsm8k} additionally evaluates the gate on a set that played no part in its construction.

\end{document}